\documentclass[conference,letterpaper]{IEEEtran}

\usepackage{fancyhdr}
\usepackage{graphicx}  
\usepackage{array}
\usepackage{fancyhdr}
\usepackage{subfigure}
\usepackage{algpseudocode}
\usepackage{algorithm}
\usepackage{bm}
\usepackage{amsmath}
\usepackage{siunitx}

\begin{document}

\title{Towards Hexapod Gait Adaptation using Enumerative Encoding of Gaits: Gradient-Free Heuristics}

\author{{Victor Parque}\\
\IEEEauthorblockA{\emph{Department of Modern Mechanical Engineering} \\ \emph{Waseda University} \\
3-4-1, Okubo, Shinjuku, Tokyo, 169-8555, Japan\\parque@aoni.waseda.jp}
}

\maketitle
\thispagestyle{plain}

\fancypagestyle{plain}{
\fancyhf{}	
\fancyfoot[L]{}
\fancyfoot[C]{}
\fancyfoot[R]{}
\renewcommand{\headrulewidth}{0pt}
\renewcommand{\footrulewidth}{0pt}
}

\pagestyle{fancy}{
\fancyhf{}
\fancyfoot[R]{}}
\renewcommand{\headrulewidth}{0pt}
\renewcommand{\footrulewidth}{0pt}

\begin{abstract}
The quest for the efficient adaptation of multilegged robotic systems to changing conditions is expected to render new insights into robotic control and locomotion. In this paper, we study the performance frontiers of the enumerative (factorial) encoding of hexapod gaits for fast recovery to conditions of leg failures. Our computational studies using five nature-inspired gradient-free optimization heuristics have shown that it is possible to render feasible recovery gait strategies that achieve minimal deviation to desired locomotion directives with a few evaluations (trials). For instance, it is possible to generate viable recovery gait strategies reaching 2.5 cm. (10 cm.) deviation on average with respect to a commanded direction with 40 - 60 (20) evaluations/trials. Our results are the potential to enable efficient adaptation to new conditions and to explore further the canonical representations for adaptation in robotic locomotion problems.
\end{abstract}

\begin{IEEEkeywords}
hexapod, gait adaptation, enumerative encoding, particle swarm optimization
\end{IEEEkeywords}

%
\IEEEpeerreviewmaketitle

\section{Introduction}

Contemporary multilegged robotic systems are expected to adapt to failing actuators on the field to continuously perform navigation, exploration, and user-defined tasks when repair is difficult\cite{taka22,joe}. In particular, robots operating in human-unfriendly environments, such as a disaster area, acquire motion by using crawling gaits or linear/rotating actuators. For instance, \emph{Quince} robot uses crawling\cite{kiri} while \emph{Spirit} uses rotating wheels. Crawling and legged mechanisms are considered more beneficial than rotating actuators because wheels are difficult to control on noisy, uneven, and steep surfaces and because crawling is more robust to failures of actuation mechanisms. Thus, crawling mechanisms are preferred in such circumstances in difficult environments.

Among the existing locomotion mechanisms, the multilegged locomotion systems have attracted the community's attention, especially quadrupeds and hexapod systems. Although the number of legs increases the complexity and weight of the overall design, the higher number of legs accomplish better crawling efficiency and better motion reliability due to redundancy of the end locomotion means. Therefore, robots with multiple leg mechanisms are the most suitable and preferable to study to be helpful in hazardous and scaffolding locations.

 \begin{figure}
 \centering
 \includegraphics[width=0.7\columnwidth]{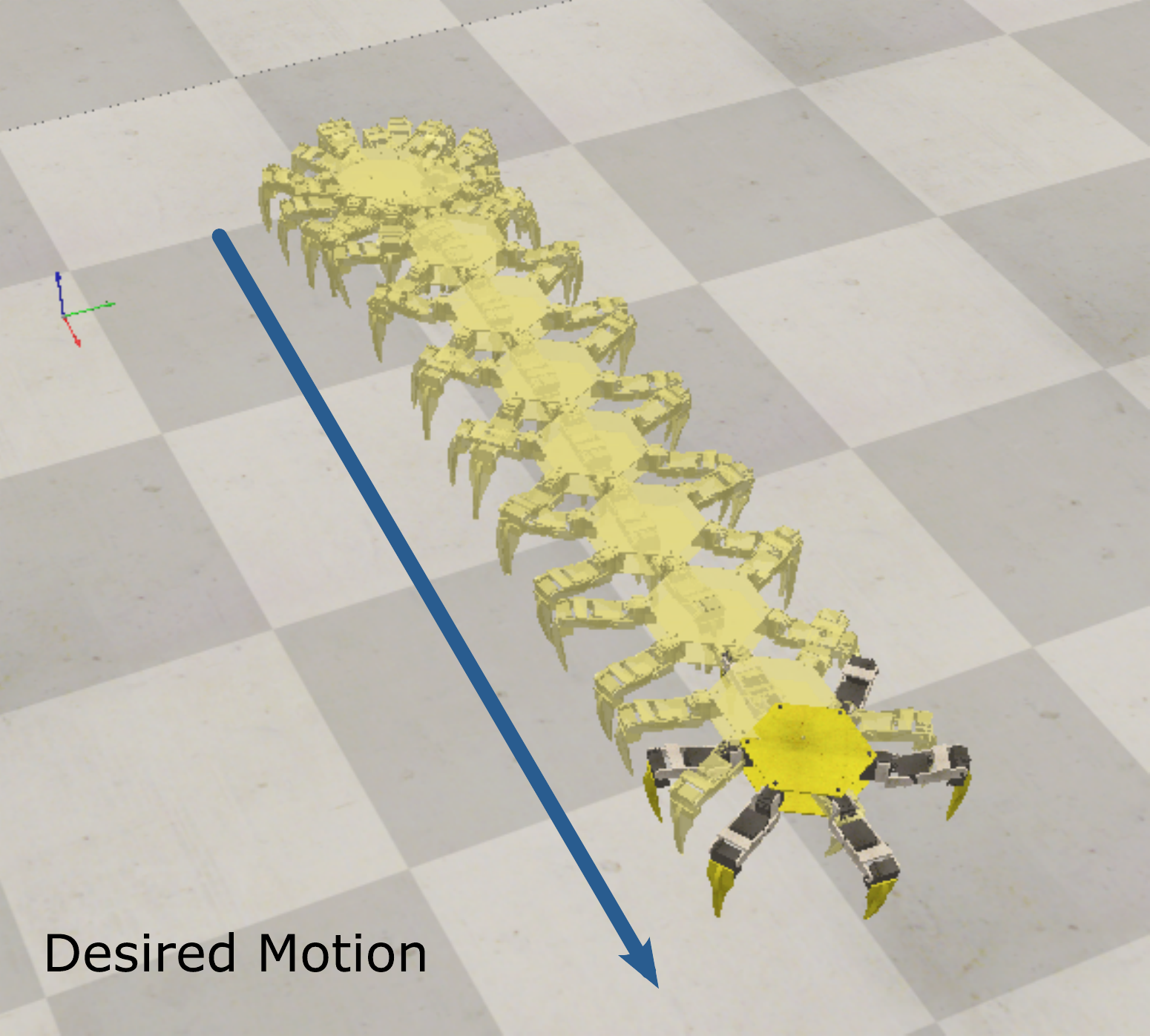}
 \caption{Example of a gait recovery of a hexapod when a leg is subject to failure. The hexapod is able to navigate in the desired direction.}
 \label{bas}
 \end{figure}

The community has focused efforts on studying Central Pattern Generators (CPG), Neural Networks (NN), and Hidden Markov Models\cite{joe} to render crawling mechanisms. CPG is inspired by natural spiking neurons to generate a walking rhythm and has been widely used to perform multilegged gaits in robots\cite{shin}. NN is a graphical/mathematical model inspired by the multiperceptron and has been widely used to generate walking gaits in legged systems\cite{shimo,suchan} as well. However, the conventional control mechanisms for legged systems suffer from adaptability issues, especially when actuators become malfunction, rendering the overall system unstable\cite{putter21}. Generally speaking, multilegged systems should adapt to malfunction quickly to enable the smooth conduction of tasks.

Furthermore, a multilegged mechanism usually responds to malfunction by selecting embedded recovery strategies or learning new recovery plans depending on its current state. Although both schemes have been studied widely during the last decade, researchers have often preferred using sample-based learning algorithms such as the ones based on gradients\cite{au69}, evolutionary computing\cite{au77,kumo,au78,au14,au73,mailer21}, Bayesian optimization\cite{cully} and reinforcement learning\cite{au76}.

The studies mentioned above rendered new learning-based adaptation strategies to respond to malfunction. For instance, recent studies have used vector-based representations of gait controllers\cite{cully,mailer21}, enabling us to understand the search and constraint space of gait adaptation mechanisms when actuators in multilegged systems become malfunction. This paper explores the benefits of using an enumerative encoding strategy to render new gait adaptation strategies when legs in hexapods malfunction. The enumerative encoding mechanism uses the factorial numbering system and uses the bijective property to the integer number system. As such, it is possible to enumerate all possible recovery gaits when legs are subject to failure. In particular, our contributions are as follows:

\begin{itemize}
\item The performance and trajectory characterization of all feasible gait strategies under a tailored duty cycle when hexapods' legs are prone to failure.
\item The study of well-known nature-inspired gradient-free optimization heuristics in tackling the gait recovery problem, showing that the gradient-free optimization heuristics are able to find recovery gaits with the small number of function evaluations.
\end{itemize}

The rest of this paper is organized as follows. Section 2 describes preliminaries in our study, section 3 describes the computational experiments, and section 4 concludes the paper.


\section{Enumerative Encoding of Gaits}

\begin{figure}
 \centering
 \includegraphics[width=0.98\columnwidth]{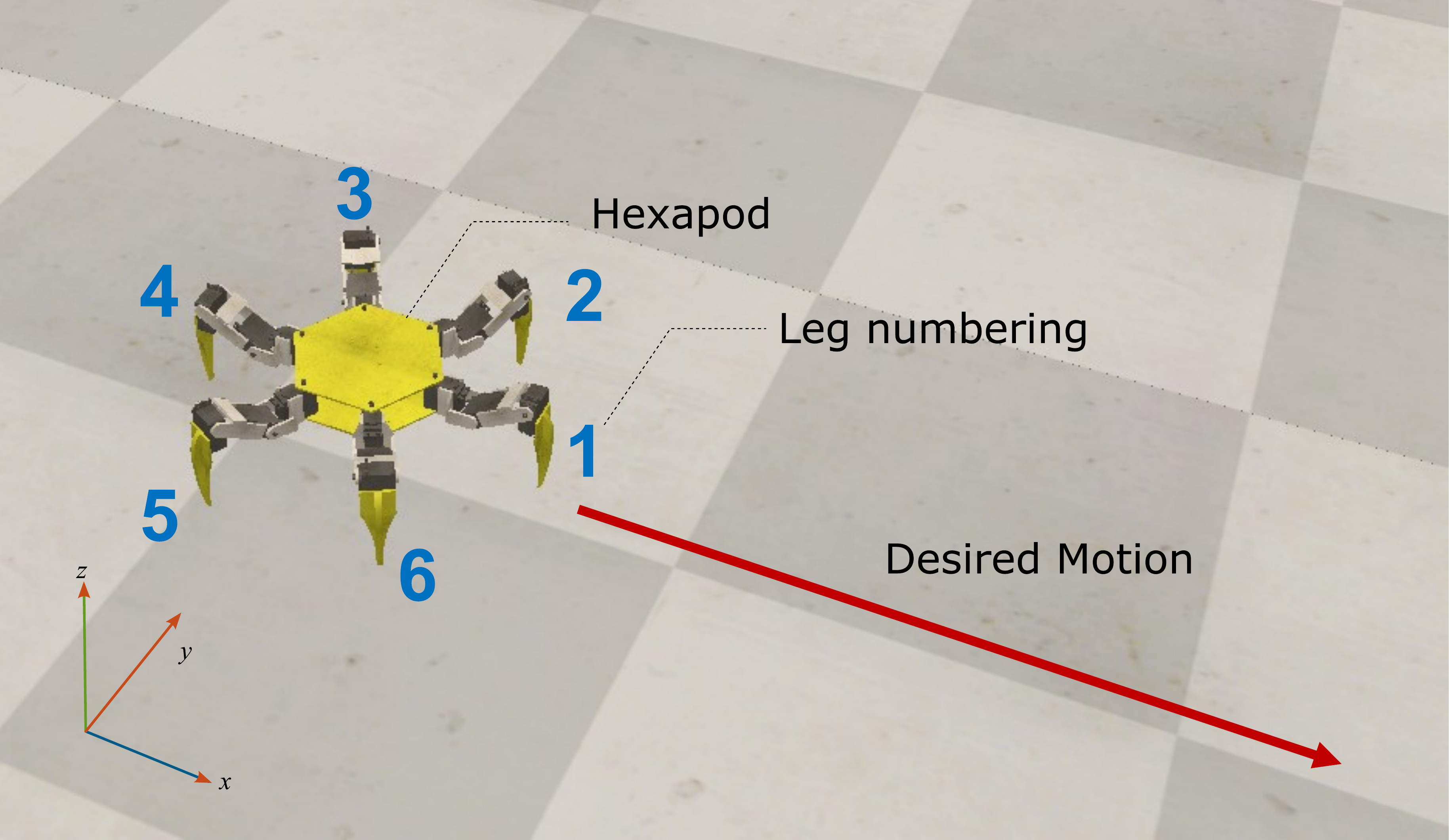}
 \caption{Hexapod model and leg numbering.}
 \label{hex}
 \end{figure}

Let a hexapod system have legs numbered with integer numbers as shown by Fig. \ref{hex}. In order to realize locomotion in the desired direction, as shown by the red line of Fig. \ref{hex}, it is essential to move the legs to maintain the desired locomotion strategy. Let the position of the $i$-th leg be represented by the vector $\bm{p}_i$, with $i \in [1, 6]$. It is possible to render gait strategies by computing offsets $\bm{d}_i$ that modify the position of each leg with respect to their initial position $\bm{p}^o_i$, and let the inverse kinematics and the control strategy to attain such offsets\cite{hex}. Thus, the position of each leg can be represented as follows:

\begin{equation}\label{eqp}
  \bm{p}_i = \bm{p}^o_i + \bm{d}_i,
\end{equation}

\begin{equation}\label{eqd}
  \bm{d}_i = (m_i(u) \cos(\theta), m_i(u) \sin(\theta), h_i(u) ),
\end{equation}
where $m_i$ is the magnitude of the offset of the $i$-th leg in the plane as a function of the normalized time $u \in [0, 1)$, $h_i$ is the height offset, and $\theta$ is the desired hexapod locomotion orientation in the plane. Naturally, $\theta \in [0^{\circ}, 360^{\circ}]$. Fig. \ref{hex} shows the direction of the desired locomotion, and considering the orientations of the x-y axis, $\theta = 0^{\circ}$ in Fig. \ref{hex}. The value of $m_i$ and $h_i$ can be computed from the normalized time $u$, as follows:

\begin{equation}\label{eqm}
m_i(u) =
\begin{cases}
    \frac{3uA}{2}   &   \text{ }  u \in [ 0, \frac{1}{3} ),    \\
    \frac{A}{2} - 3A(u-\frac{1}{3})   &   \text{ } u \in [\frac{1}{3}, \frac{1}{2}), \\
    - 3A(u-\frac{1}{2})   &   \text{ } u \in [\frac{1}{2}, \frac{2}{3}), \\
    -\frac{A}{2} \Big (1 - 3(u - \frac{2}{3}) \Big)    &   \text{ } u \in [\frac{2}{3}, 1),
\end{cases}
\end{equation}

\begin{equation}\label{eqh}
h_i(u) =
\begin{cases}
    0   &   \text{ }  u \in [ 0, \frac{1}{3} ),    \\
    6H(u-\frac{1}{3})   &   \text{ } u \in [\frac{1}{3}, \frac{1}{2}), \\
    H\Big ( 1 - 6(u-\frac{1}{2}) \Big )  &   \text{ } u \in [\frac{1}{2}, \frac{2}{3}), \\
    0   &   \text{ } u \in [\frac{2}{3}, 1),
\end{cases}
\end{equation}

In the above-mentioned formulations, the constant $A$ refers to the step amplitude, and $H$ refers to the step height, both of which can be set according to the size of the hexapod or structure of the legs. Also, the normalized time $u$ is a function of the duty cycle of the $i$-th leg, i.e., it is dependent on the actual or simulation time and the order in which each leg is used. Then, for each leg $i\ \in [1, 6]$:

\begin{equation}\label{eqd}
u = \mod \Big ( t + \frac{g_i - 1}{6}, 1 \Big ),
\end{equation}
where $t$ is the actual or simulation time and $g_i \in [1, 6]$ is the index of the $i$-th leg used for movement. The division by six is due to the six legs of the hexapod system, and the modulus operator is to ensure that $u \in [0, 1)$ is suitably normalized. As such, the configuration of $g_i$ is basically the permutation of legs that enable the above-mentioned gait strategy and can be encoded by the factorial numbering system:

\begin{equation}\label{fac}
  g = ( g_1,  g_2,  ..., g_6  ),
\end{equation}
where $g_i$ is the index of the leg used for locomotion in the $i$-th order. For instance, $g = (5, 1, 2, 4, 3, 6)$ represents a gait strategy in which leg 5, leg 1, leg 2, leg 4, leg 3 and leg 6 are used in such order to compute the offsets for legs 1...6. Naturally, the above representation is a subset of all possible hexapod gait encodings, and since the factorial number system is used, it is possible to render all possible gaits from the 1-1 bijection to the integer set $\{1, ...,n! \}$.

Now, when a leg in a hexapod system malfunctions, it is expected that the remaining five legs will render a limited set of gait strategies (5! = 120), for which it is possible to study gait adaptations for particular friction parameters in the actuators and the floor.

Compared to conventional vector-based encodings\cite{mailer21}, the factorial representation brings the potential of introducing \emph{canonicity} and \emph{diversity} concepts into the gradient-free optimization algorithms.

\begin{itemize}
  \item \emph{Canonicity} implies that every solution in the search space can be rendered from a unique and single encoding, without repetitions, and while avoiding potential repair operators. Although this notion was introduced in the context of Genetic Programming\cite{cangp}, the idea of allowing unique representations enables to construct a search space that is convex, in principle.

  \item Since each representation in the factorial search space is unique, it becomes potential to sample previously unsampled solutions since the distance metrics can be computed from the difference between two numbers. In the context of a hexapod system, since the factorial number system has a 1-1 bijection to integer numbers, it is possible to search in the space of numbers to explore qualitatively different hexapod gaits as two different integer numbers will represent behaviourally different locomotion strategies.

  \item Also, since gait strategies can be encoded by integer numbers, then it is possible to use parallel computing schemes to explore the performance of behaviourally distinct gait strategies in tandem.
\end{itemize}

 \begin{figure*}[ht]
 \centering
 \includegraphics[width=1\textwidth]{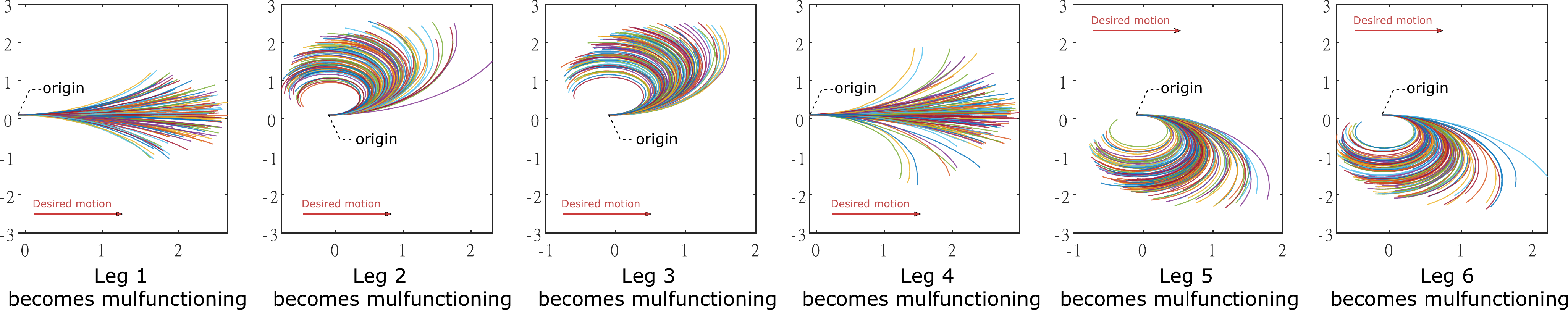}
 \caption{Trajectories of the hexapod when a leg's malfunction occurs. Each figure shows all possible trajectories when the hexapod uses a unique gait encoded by the factorial number system under a specific occurrence of leg malfunction. On the left, the \emph{origin} denotes the start position of the hexapod, and each curve represents the path traveled by the robot.}
 \label{fixLeg}
 \end{figure*}

 \begin{figure*}[ht]
 \centering
 \includegraphics[width=1\textwidth]{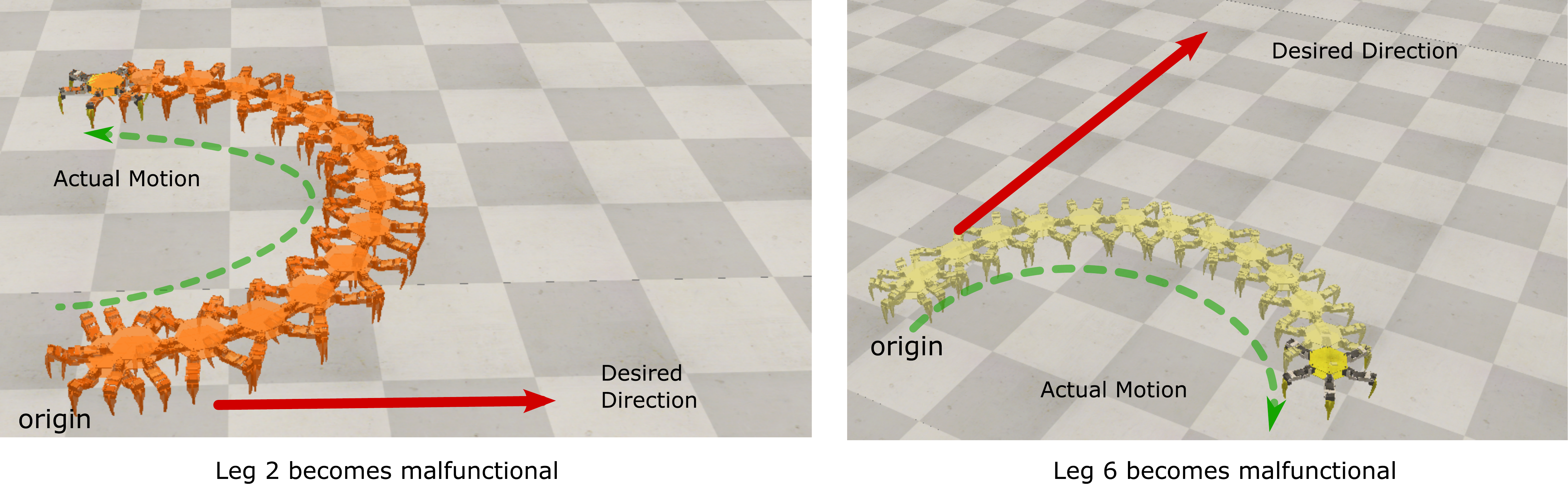}
 \caption{Example of trajectories of the hexapod when leg's malfunction occurs.}
 \label{fixLegSample}
 \end{figure*}

\section{Computational Experiments}

In order to simulate the behavior of a hexapod robot under dynamic considerations, we used the Coppelia Simulation software \cite{hex}. Fig. \ref{hex} shows the structure of the robot used in our study, for which the hexapod is rotationally symmetric, each of the legs has three degrees of freedom, and the legs are labeled with an index, such as 1, 2, ..., 6.

\begin{itemize}
  \item we studied leg failures in the hexapod system and explored possible recovery gait configurations to move according to a user-defined orientation under the enumerative encoding strategy presented in the previous section.
  \item we used gradient-free optimization heuristics to study the feasibility of rendering gait recoveries with a low number of fitness function evaluations.
\end{itemize}

For simplicity and without loss of generality, we set the hexapod system to follow according to a desired orientation in the plane as shown by the red line in Fig. \ref{hex}, that is $\theta = 0^{\circ}$. Realizing the locomotion strategies in different orientations is straightforward. Studying the trajectory tracking for non-linear paths is out of the scope of this paper. Also, we let the step amplitude $A = 0.11E$, and the step height $H = 0.02E$  be fractions of the overall size factor $E$ or the hexapod\cite{hex}.

Furthermore, our computing environment consisted of an Intel i9-9900K, 3.6GHz, and our gradient-free optimization heuristics were implemented in Matlab 2021b. The Matlab-Coppelia interface was based on WebSockets.

\subsection{Study of Leg's Malfunction}

In this section, we studied the trajectory behavior of the overall system when the hexapod's legs malfunctioned and when the hexapod used all possible gait strategies under the enumerative (factorial) encoding. We investigated the trajectories when the hexapod is requested to navigate according to the user-defined orientation $\theta = 0^{\circ}$ in the plane and when each particular legs malfunction. Here,

\begin{enumerate}
  \item When a leg becomes malfunctioning, five legs remain functioning, and we generated all 5! = 120 gait recoveries.
  \item Given a gait strategy, we let the hexapod navigate towards the user-defined orientation $\theta = 0^{\circ}$ during 30 s.
  \item We logged the trajectory for each case of combining a malfunctioning leg and a gait recovery strategy.
\end{enumerate}

Although it is possible to skip and lock the malfunctioning leg to avoid contact with the floor, as the related works\cite{mailer21}, we rather let the malfunctioning legs contact the floor to simulate the challenging conditions of navigating under friction and varying overall system dynamics.

To show the performance of the overall system, Fig. \ref{fixLeg} shows the trajectories of the hexapod when particular legs malfunction, and Fig. \ref{fixLegSample} shows examples of the trajectory followed by the hexapod using an arbitrary gait strategy when leg 2 and leg 6 became malfunctioning. By observing the trajectories from Fig. \ref{fixLeg} and Fig. \ref{fixLegSample}, we can note the following facts:

\begin{itemize}
  \item Each sub-figure of Fig. \ref{fixLeg} shows all the origin (start) of the location of the hexapod system and the followed trajectories when the hexapod used all computable gait strategies from the enumerative (factorial) encoding system under each scenario of leg failure.
  \item For instance, Fig. \ref{fixLeg} shows that when leg 2 and leg 3 malfunction, the hexapod tends to move towards the left-side in counterclockwise fashion.
  \item On the other hand, when leg 5 and leg 6 become malfunction, the hexapod tends to move towards the right side in a clock-wise fashion. This phenomenon is due to leg 2/leg 3 (leg 5/leg 6) being located on the left (right) side of the plane that divides the hexapod's x-axis, as shown in Fig. \ref{hex}; thus, the weight and friction with the floor make the system relatively unstable.
  \item Furthermore, when leg 1 and leg 4 become malfunction, the hexapod either follows a relatively straight trajectory or tends to turn left or right, depending on the kind of gait strategy used.
\end{itemize}

The results mentioned in this section pinpoint the feasibility of computing gait recovery strategies under the enumerative (factorial) encoding scheme. By exploring the performance of trajectories with respect to deviation to a line, Fig. \ref{fixLeg} can be used to consider the alternative gait recovery plans that best meet the desired movement direction and goal.

\subsection{Gradient-Free Heuristics}

The previous section has shown the performance of the overall hexapod system in situations of leg failures. In changing conditions of floor dynamics and online performance considerations, it would be desirable to render recovery strategies under a small number of function evaluations to meet the locomotion in the desired direction.

Since the gait strategies can be mapped into the integer system, it becomes possible to search in the space of numbers with a lower bound set as one and the upper bound set as 5! = 120 (assuming one leg is subject to failure). Naturally, it will be necessary to use a gradient-based optimization algorithm to assess the integer-based search space. Here,

\begin{enumerate}
  \item For each scenario of leg failure and each algorithm, the objective function $F$ is set as the deviation in the commanded direction.
  \item As such, the goal of the optimization heuristic will be to compute a gait recovery plan that allows the hexapod to navigate along the line of the x-axis (i.e., the orientation $\theta = 0^{\circ}$) during 30 s.
\end{enumerate}

\begin{figure}
 \centering
 \includegraphics[width=1\columnwidth]{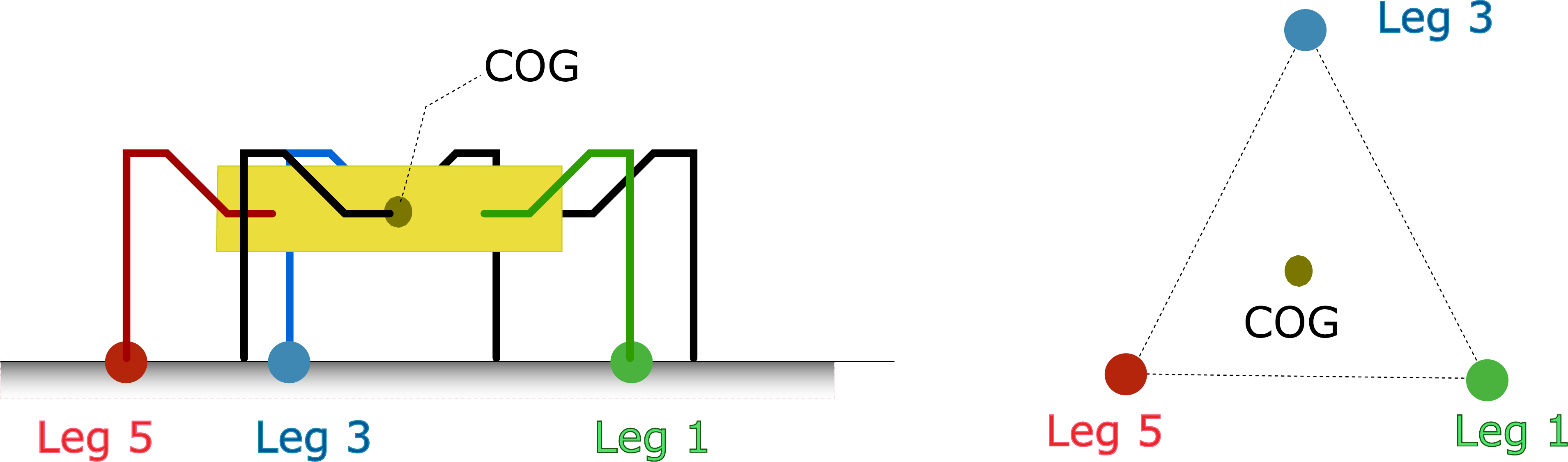}
 \caption{Basic concept of stability.}
 \label{stabi}
 \end{figure}

Also, before locomotion, we allow the robot to rotate in its own axis to self-adjust the balance of the overall robot system and to meet the desired locomotion in the user-defined orientation $\theta = 0^{\circ}$. This strategy allows the system to solve the stability problem first when the leg touches the floor and to avoid sudden turns to the right or the left side, as shown in Fig. \ref{fixLeg}, due to the unstable distribution of weight. In principle, by rotating on its own axis, the hexapod aims at finding the balanced distribution of the remaining legs contacting the floor. One wishes that the Centre of Gravity (COG) be within the convex hull of the projection of the legs at the floor, as shown by Fig. \ref{stabi}. Then, the role of the optimization heuristic is to find the suitable gait strategy to achieve minimal deviation to the desired orientation in locomotion. The reader may note that by tackling this goal, we also implicitly tackle the dynamic balancing of the system.

  \begin{figure*}[ht]
 \centering
 \includegraphics[width=1\textwidth]{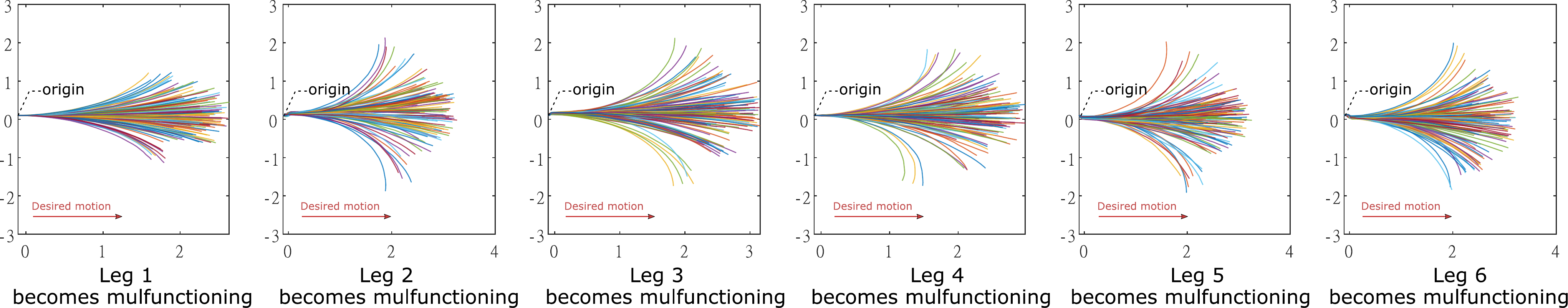}
 \caption{Effect of self-adjusting the orientation of hexapod in its own axis. Trajectories of the hexapod when a leg's malfunction occurs, considering a rotation on its own axis. Each figure shows the trajectories when the hexapod uses a unique gait encoded by the factorial number system. On the left, the \emph{origin} denotes the start position of the hexapod, and each curve represents the path traveled by the robot.}
 \label{fixRotlegs}
 \end{figure*}

\begin{figure*}[ht!]
	\begin{center}
		\subfigure[Leg 1 becomes malfunctioning]{\includegraphics[width=0.325\textwidth]{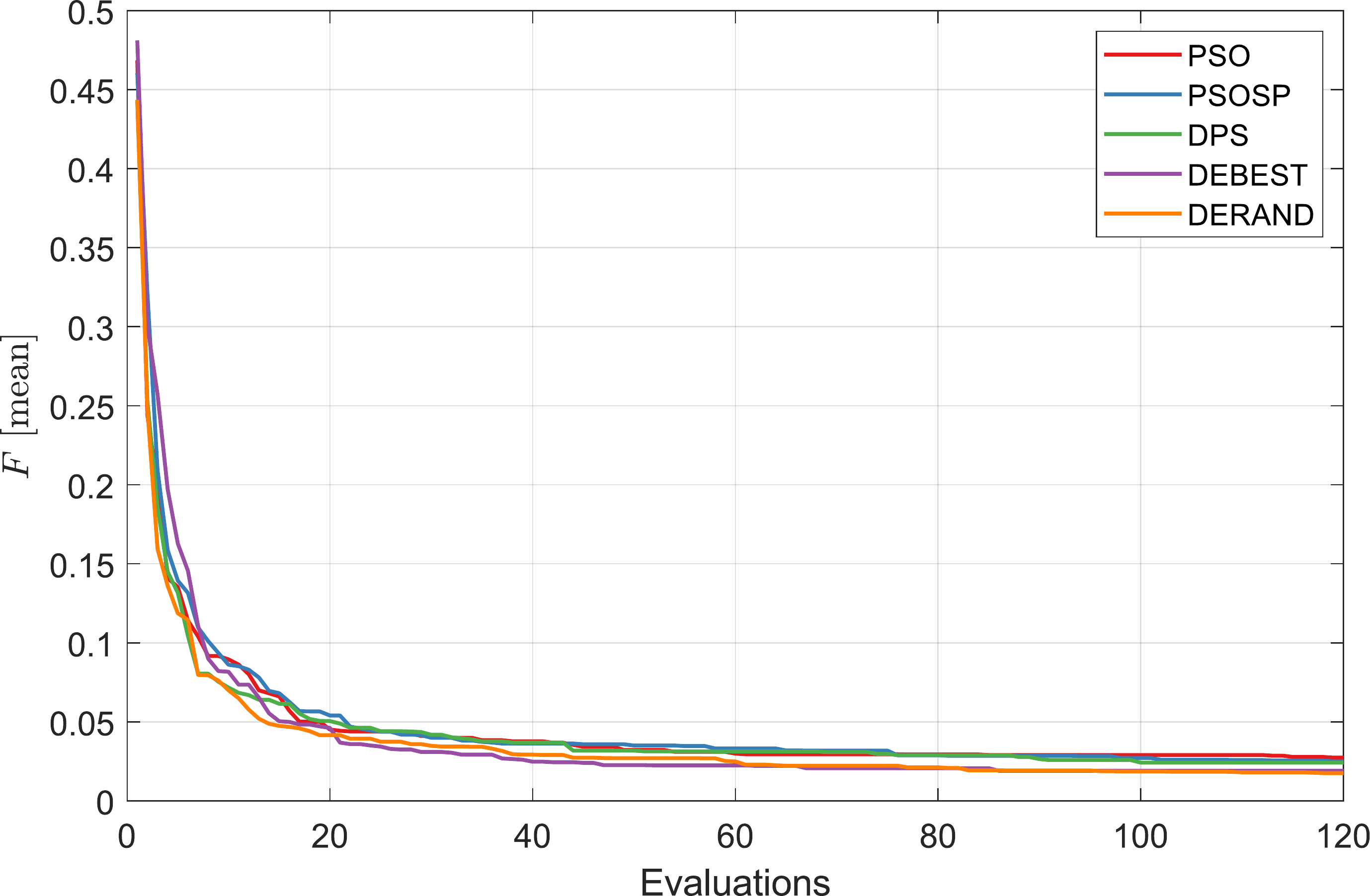}}
		\hfill
		\subfigure[Leg 2 becomes malfunctioning]{\includegraphics[width=0.325\textwidth]{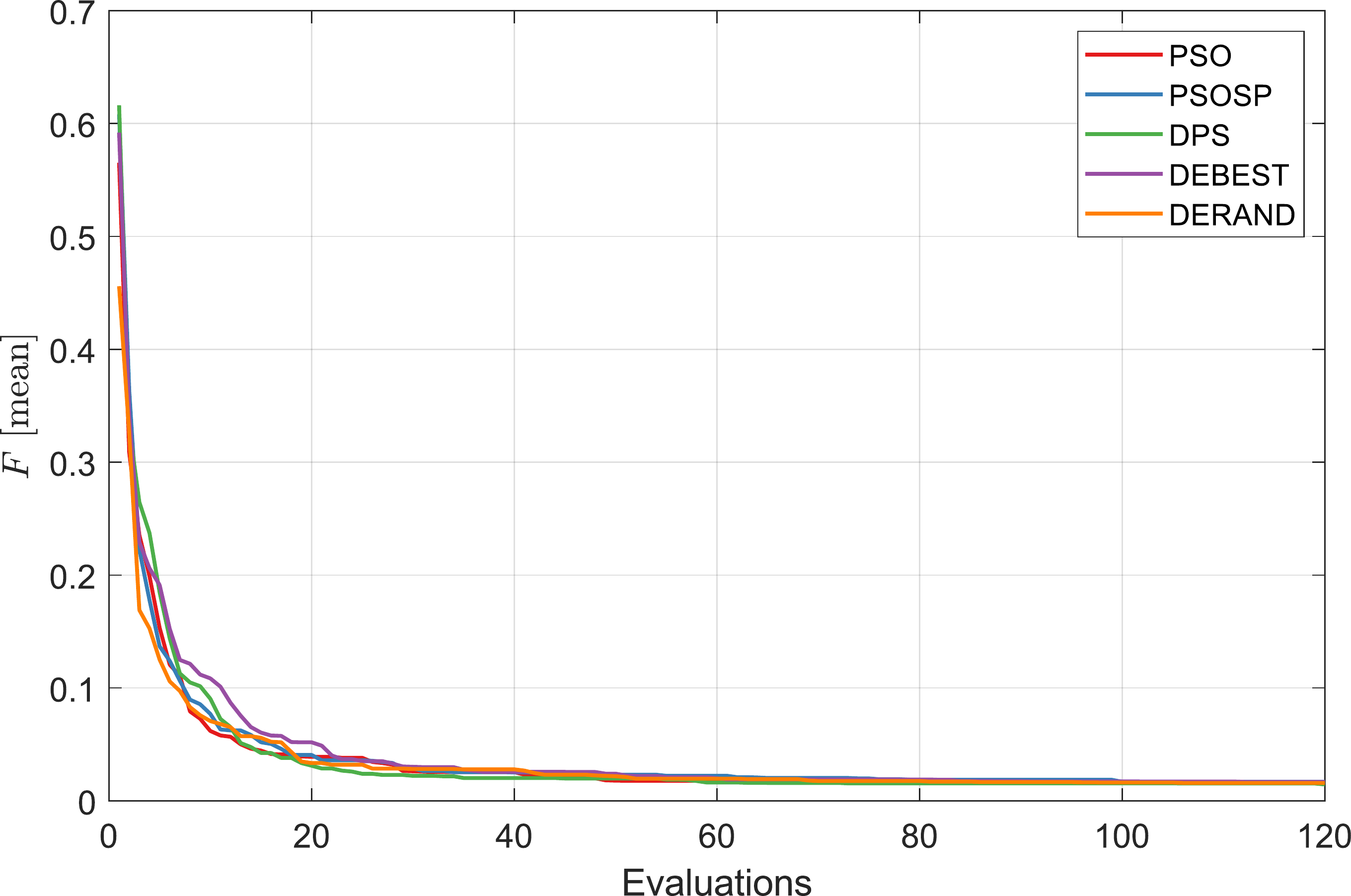}}
		\hfill
		\subfigure[Leg 3 becomes malfunctioning]{\includegraphics[width=0.325\textwidth]{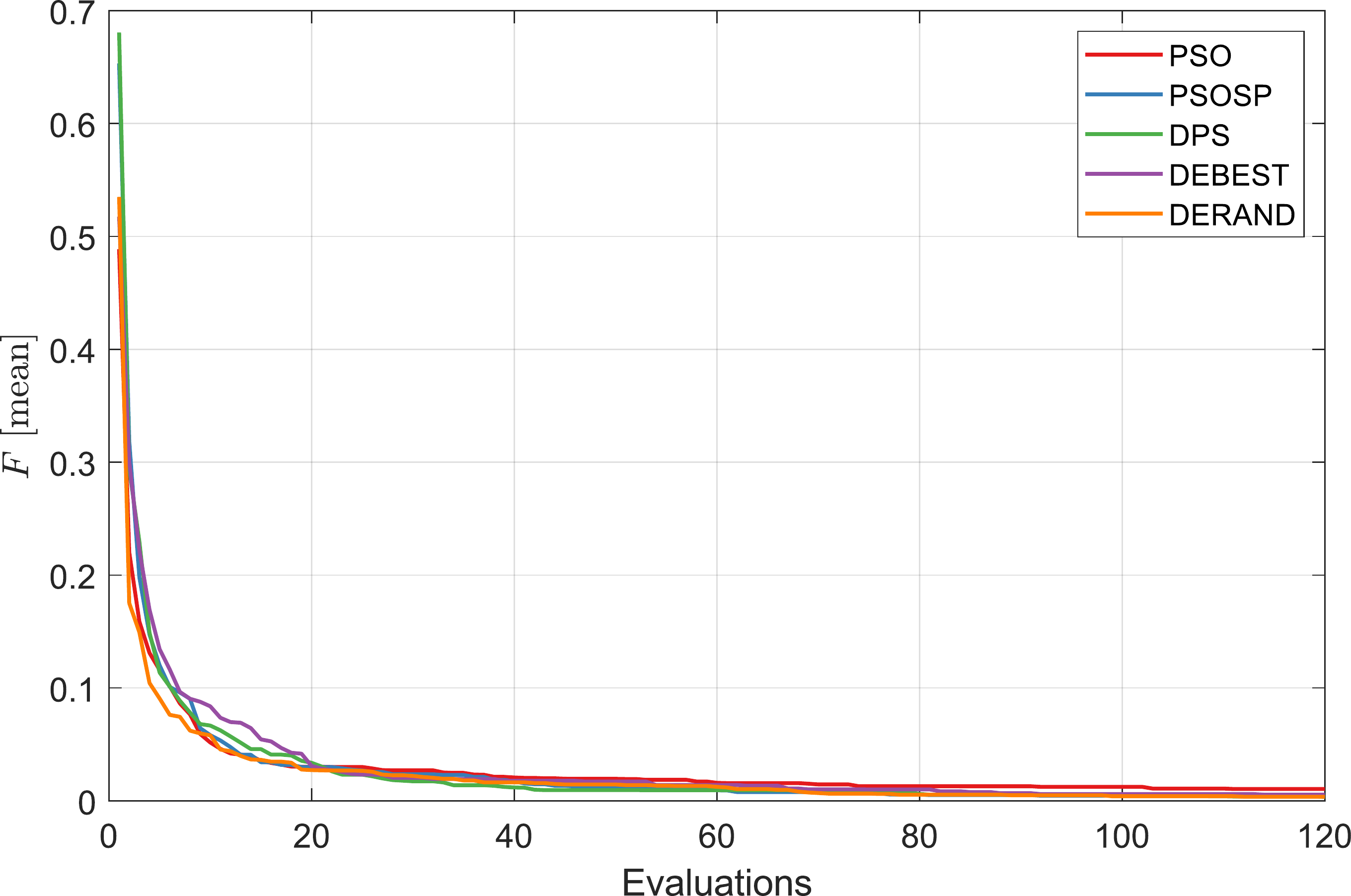}}
		\hfill
		\subfigure[Leg 4 becomes malfunctioning]{\includegraphics[width=0.325\textwidth]{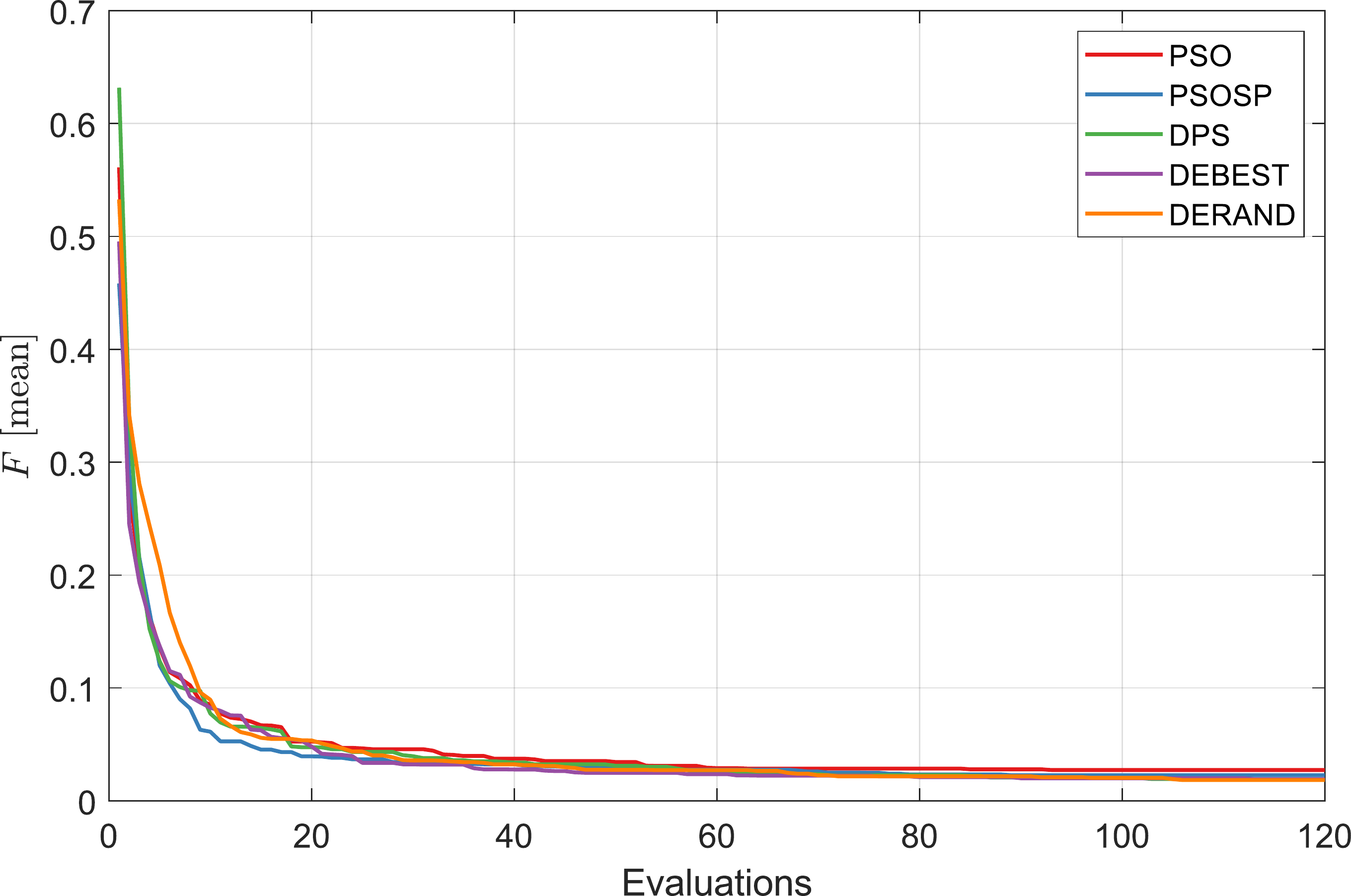}}
		\hfill
		\subfigure[Leg 5 becomes malfunctioning]{\includegraphics[width=0.325\textwidth]{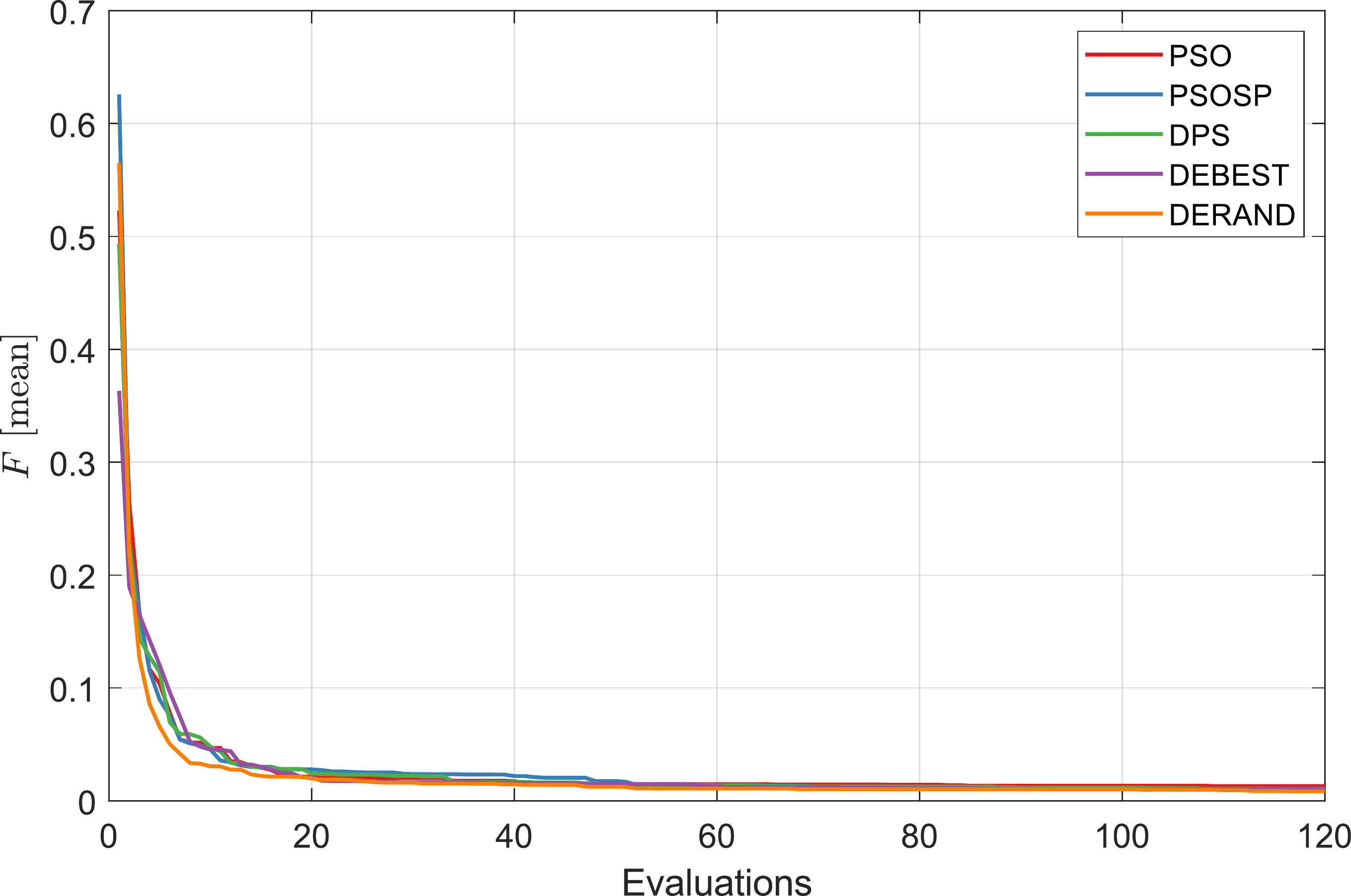}}
		\hfill
		\subfigure[Leg 6 becomes malfunctioning]{\includegraphics[width=0.325\textwidth]{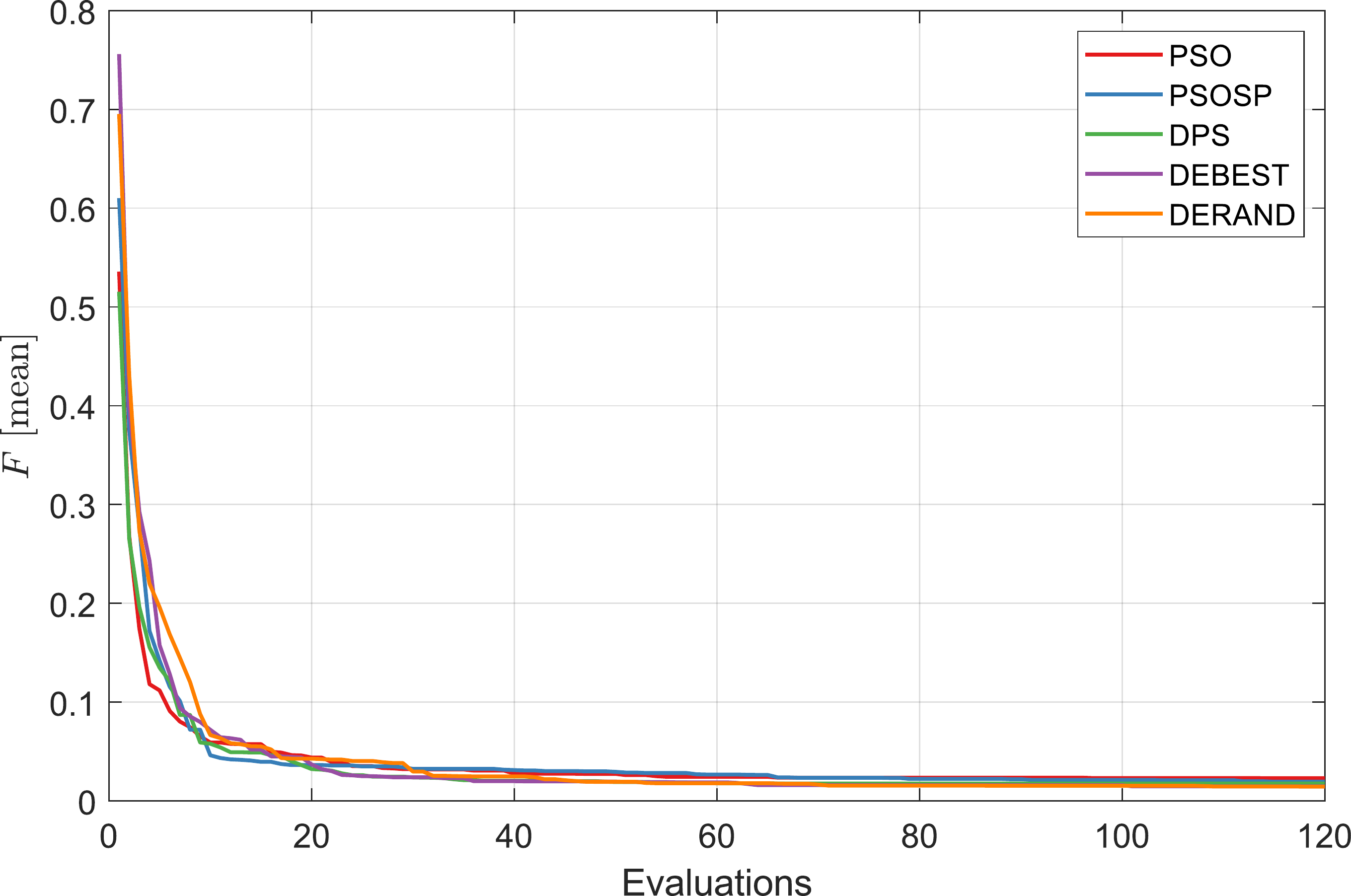}}
		\hfill
	\end{center}
	\caption{Convergence of the fitness function for minimal deviation along a line. The units of the objective function are in meters, thus smaller absolute deviation to the goal is desirable. The convergence curves represent the mean over 30 independent runs when the hexapod has a leg under failure condition.}
	\label{conv}
\end{figure*}

 \begin{figure*}
 \centering
 \includegraphics[width=0.98\textwidth]{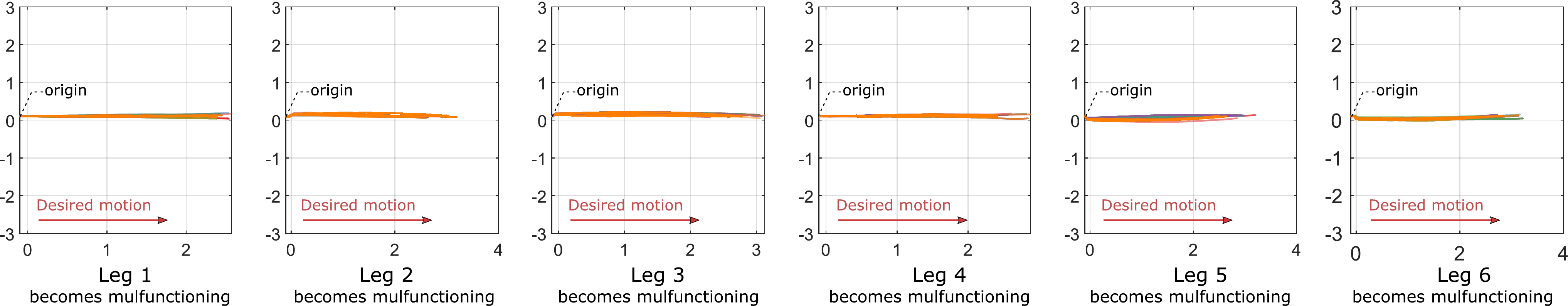}
 \caption{Best trajectories found by the optimization heuristics over independent runs.}
 \label{paths}
 \end{figure*}

 \begin{figure*}
 \centering
 \includegraphics[width=0.98\textwidth]{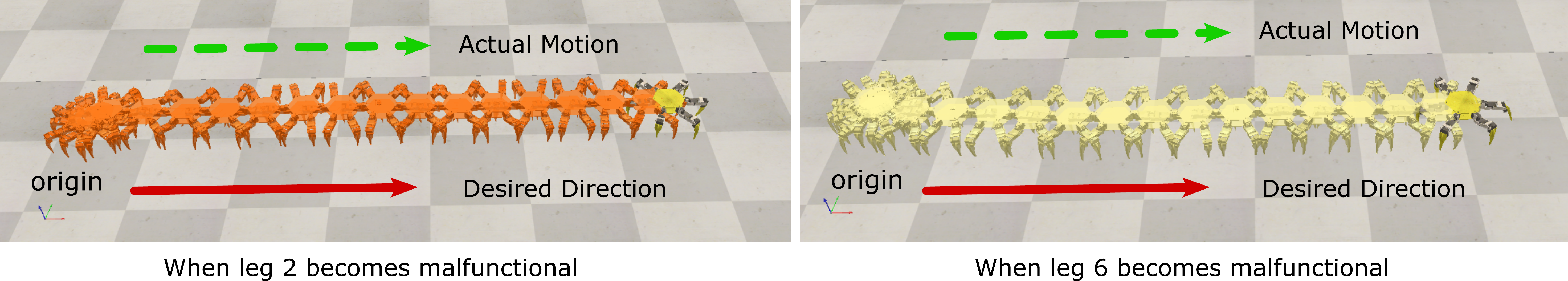}
 \caption{Example of trajectories found by the optimization heuristics.}
 \label{pathSample}
 \end{figure*}

This subsection studies the performance of well-known nature-inspired optimization heuristics to attain effective gait recoveries. As such, we studies the performance of Particle Swarm Optimization (PSO)\cite{pso95}, Particle Swarm Optimization with Speciation (PSOSP)\cite{psosp}, Differential Particle Scheme (DPS)\cite{dps21}, Differential Evolution with BEST/1/BIN mutation (DEBEST)\cite{de97}, Differential Evolution with RAND/1/BIN mutation (DERAND)\cite{de97}. Parameters for algorithms based on Particle Swarms involve $\omega$  =  0.5, weight on pbest $c_1$  =  0.5, weight on gnest $c_2$  =  2, population size was set as 10. Other parameters implement the default configurations stated in the references. As for Differential Evolution algorithms, the crossover rate was $CR = 0.5$, the scaling factor $F = 0.7$. The fine tuning of the above-mentioned parameters is out of the scope of this paper.

For evaluations, we used 30 independent runs with 120 function evaluations at the maximum (since 5! = 120 corresponds to the maximum number of gait strategies when a leg is subject to failure), enabling the comparison of various modes of exploration/exploitation under tight computational budgets.

In order to show the nature of the search space under the above-mentioned considerations, Fig. \ref{fixRotlegs} shows the trajectories of the hexapod system for different scenarios of leg failure. By observing the results in Fig. \ref{fixRotlegs}, we note the following facts:

\begin{itemize}
  \item Each sub-figure shows the origin (start) and the followed trajectories for each condition of leg failure and each gait rendered from the enumerative (factorial) encoding scheme.
  \item Each sub-figure of Fig. \ref{fixRotlegs} shows the effect of self-rotation in the own axis is able to solve the balancing of the overall system and to avoid motions towards undesirable directions (i.e., towards the left or right side as shown by Fig. \ref{fixRotlegs}).
  \item Although each sub-figure shows that certain gait strategies make the hexapod turn towards the left or right side, Fig. \ref{fixRotlegs}) also shows that there exists a number of gait strategies that achieve small deviation with respect to the x-axis (i.e., the desired orientation of the locomotion).
\end{itemize}

%

In order to show the behavior of the convergence algorithm across the 30 independent runs, Fig. \ref{conv} shows the number of function evaluations on the x-axis, and the mean convergence of the function $F$, in the y-axis (units in \si{\m}). By observing Fig. \ref{conv}, we can note the following facts:

\begin{itemize}
  \item The fitness values converge to values around 2.5 \si{\cm} in most of the cases for each situation of leg failure. These results imply that the optimization heuristics were able to generate hexapod gaits that achieved minimal deviation.
  \item The fastest convergence occurs in the period of less than 0-20 function evaluations, whereas the convergence becomes relatively flat over the region 20-120. This observation implies that it becomes relatively simple and tractable to generate gait recoveries that aim to achieve locomotion in a desired direction.
  \item The convergence to minimal values occurs in about 40-60 evaluations. Also, it is possible to converge to a deviation of about 10 \si{\cm} in less than 20 function evaluations. This observation shows that it is possible to obtain feasible and effective recovery gaits with a small number of function evaluations, implying that a large number of simulations and real-world trials may not be necessary as suggested by the related works\cite{cully,mailer21}.
  \item All studied algorithms were able to attain competitive performance under a small number of evaluations, implying that the search space is amenable to the characteristics of the studied heuristics. Studying the performance of different and tailored search heuristics is out of the scope of this paper and left for future work.
\end{itemize}

To portray the types of gait recovery plans that our approach was able to generate, Fig. \ref{paths} shows the trajectories of the hexapod of all algorithms when particular legs become malfunctioning. Basically, Fig. \ref{paths} shows the trajectories of all gait recoveries obtained from all algorithms' overall independent runs. Also, Fig. \ref{pathSample} shows samples of trajectories when the hexapod has leg 2 and leg 6 under failure conditions. By observing Fig. \ref{paths} and Fig. \ref{pathSample}, we can observe the following:

\begin{itemize}
  \item The hexapod was able to navigate while meeting the desired orientation $\theta = 0^{\circ}$ reasonably well for each situation of leg failure.
  \item All optimization heuristics rendered reasonably good gait strategies that meet the navigation requirements.
  \item Compared to Fig. \ref{fixLegSample}, the trajectories shown in Fig. \ref{pathSample} show that it is possible to overcome the undesirable pattern to have the system moving to the left and right sides.
  \end{itemize}

The above-mentioned observations pinpoint the potential of using optimization heuristics to generate gait recoveries that attain minimal deviation to locomotion directives. In future works, we plan to explore the role of synchrony and parallelism in rendering feasible gaits\cite{iconip14}, as well study the role of combinatorial and graph structures to consider the motion of multiple legs at the time\cite{gecco21}. Also, we plan to study whether a tailored class of optimization heuristics would solve the gait recovery under the enumerative encoding scheme.



\section{Conclusion}

In this paper, we have studied the feasibility of using an enumerative (factorial) encoding for hexapod gait recovery to conditions of leg failures. We also have evaluated the extent and degree of preserving the locomotion directives while exploring all possible gait strategies. Furthermore, we have evaluated the performance of five nature-inspired gradient-free optimization heuristics to render effective gait recovery strategies. Our computational studies have shown that it is possible to render feasible recovery gait strategies that achieve minimal deviation to desired locomotion directives in a few evaluations (trials). In particular, we found that it is possible to render gait strategies achieving 2.5 \si{\cm} (10 \si{\cm} ) error in average with respect to a commanded direction with 40-60 (20) evaluations/trials. In future work, we plan to study the performance of parallelism, combinations, and tailored heuristics for gait adaptation. We argue that using canonical encoding mechanisms\cite{cangp} is beneficial to allowing efficient fault-tolerant and adaptive locomotion systems.

\section*{Acknowledgement}

This research was supported by JSPS KAKENHI Grant Number 20K11998.

\end{document}